\documentclass{article}
\usepackage{arxiv}

\usepackage{amsthm}
\usepackage{multirow}
\usepackage{subcaption}
\usepackage{mathtools}
\usepackage[export]{adjustbox}

\usepackage[utf8]{inputenc} 
\usepackage[T1]{fontenc}    
\usepackage{hyperref}       
\usepackage{url}            
\usepackage{booktabs}       
\usepackage{amsfonts}       
\usepackage{nicefrac}       

\newtheorem{theorem}{Theorem}[section]

\newtheorem{lemma}[theorem]{Lemma}

\title{Quantifying Challenges in the Application of Graph Representation Learning}

\author{Antonia Gogoglou, C. Bayan Bruss, Brian Nguyen, Reza Sarshogh, Keegan E. Hines \\
Capital One,
McLean, VA 22102}

\begin{document}
\maketitle

\begin{abstract}
{\it Graph Representation Learning (GRL)} has experienced significant progress as a means to extract structural information in a meaningful way for subsequent learning tasks. Current approaches including {\it shallow embeddings} and {\it Graph Neural Networks} have mostly been tested with node classification and link prediction tasks. In this work, we provide an application oriented perspective to a set of popular embedding approaches and evaluate their representational power with respect to real-world graph properties. We implement an extensive empirical data-driven framework to challenge existing norms regarding the expressive power of embedding approaches in graphs with varying patterns along with a theoretical analysis of the limitations we discovered in this process. Our results suggest that "one-to-fit-all" GRL approaches are hard to define in real-world scenarios and as new methods are being introduced they should be explicit about their ability to capture graph properties and their applicability in datasets with non-trivial structural differences. 
\end{abstract}


\section{Introduction}
In recent years, {\it Graph Representation Learning} (GRL) has emerged as a promising tool to utilize the rich information that lies in complex network structures. The effectiveness of these methods has led to them finding applications in molecular physics \cite{pmlrSanchez2018}, biology \cite{battaglia2018relational}, e-commerce \cite{wang2018billion}, online platforms \cite{grbovic2018real} and finance \cite{bruss2019DeepTrax, bruss2019Scale}. 
Viewed under a unified prism, GRL methods are trained to embed closer together the nodes that appear in the same {\it context} as one another \cite{Khosla2019Comparative}. However, the manner in which they define this context varies significantly affecting both scalability and performance. Its purest form would be the full adjacency matrix, while other more computationally efficient approximations include defining a sample of the context either through random walks or subsampling of neighborhoods \cite{marsden2013eigenvalues}. Amongst deep learning approaches in this field, Graph Convolutional Networks (GCNs) have gained in popularity due to their smaller number of trainable parameters, their ability to handle arbitrary network geometries and produce relational information in an inductive way. Variations of GCNs have also been employed in a unified training setting to extract information from the graph topology while optimizing for a set of node labels. 



Traditional graph mining techniques, the predecessors of GRL, were evaluated on their ability to express properties of the original graph, e.g. community structure, degree distribution, centralities and role preservation to name a few \cite{Faloutsos2006Sample, henderson2012rolx}. Similarly, the effectiveness of GRL methods should be tested with regard to these complex graph characteristics \cite{Dalmia2018WWW}. It is stated in \cite{dehmamy2019} that representing a wide range of graph statistics is a non trivial task for GCNs. Even though some of the theoretical limitations of GCNs have been explored in recent studies, a rigorous data-driven comparison of approaches in this regard has not been conducted yet. 

Recent concerns in the GRL community regarding the lack of reliable benchmarks to evaluate these methods have been raised leading to the creation of the Open Graph Benchmark initiative \footnote{https://ogb.stanford.edu/}.
In the state of the art, each new approach focuses on slightly improving accuracy in either node classification or link prediction tasks often based on small datasets with non complex characteristics. Moreover, assumptions about the data remain unaddressed with some examples including the density and transitivity of the original graph, the use of the strongest connected components in the training process or the effects of uniform sampling on the degree distribution. However, selecting one of the state of the art methods for a real-world application scenario could depend heavily on these assumptions.

In this work, we utilize a rigorous data driven framework to discover the limitations and possible obstacles in the application of popular GRL approaches. We provide both theoretical and experimental evidence on the relationship between characteristics of the graph, the considered methods and their hyperparameters. Our objective is to shed light on the context GRL methods extract from a graph and the properties that are preserved in the latent space without focusing on optimizing the inference process as happens when representation learning is conditioned upon a specific set of labels.
Unified training of representation learning with node classification tasks is out of scope in this analysis since the objective in those tasks would be the inference of the particular label instead of the accuracy of the representation. Moreover, unified training is not always a valid option in real-world scenarios with high class imbalance, but enhancement in performance can still be observed by incorporating the appropriate topological information that best fits the properties of the dataset at hand. 

Our contributions:
\begin{itemize}
    \item We select a wide range of state of the art GRL methods to be tested in an unsupervised embedding generation setting. We utilize open source datasets commonly used in literature with the addition of synthetically generated ones with data properties similar to what is often exhibited by real world graphs.
    
    \item We provide theoretical proofs of the relationship of GRL methods to graph properties, including degree, community memberships and triangle counts. 
    
    \item Going beyond this theoretical background, we design an evaluation framework that challenges unsupervised GRL approaches in their ability to encapsulate information about properties of the original graph and connect them to the applicability of each method dependent on structural complexities that need to be preserved in the embedding space.
    
    \item We then investigate these principles in practice and conduct an extensive sensitivity analysis on the interaction between method selection, hyperparameters and data characteristics for any practitioner interested in deploying GRL methods. We connect the applicability of the methods to the type of graph, its density and connectivity patterns and suggest that there is no "one-to-fit-all" approaches in GRL. 
\end{itemize}

\section{Related Work}
Graph representation learning seeks to define a function that can map a geometric graph structure to a low-dimensional, most commonly, euclidean space. This function is usually parameterized to learn or approximate some correspondence of the geometric distance between nodes to distance in the desired continuous space.  Early efforts focused on factorization approaches such as Singular Value Decomposition or computing the eigenvectors of the graph Laplacian \cite{belkin2003laplacian} \cite{tenenbaum2000global} \cite{roweis2000nonlinear}. These efforts built on an even longer history of spectral graph theory which had been studying the relationships between the eigenvalues of a graph Laplacian and its connectedness \cite{marsden2013eigenvalues}. One of the big challenges to these methods is applicability on very large graphs.

Subsequent GRL efforts found scalable approaches for large natural graphs by applying effective word embedding techniques from NLP to graphs. Sequences of nodes are generated via random, or biased, walks on the graph which are then fed into shallow single layer skip-gram models \cite{perozzi2014deepwalk}, \cite{grover2016node2vec}. The random walks can be viewed as a type of sampling from a distribution of pairwise proximities between all nodes. The skip-gram model implicitly learns to map nodes in those samples to Euclidean space preserving neighborhood structure. Most recently, efforts to connect graph learning to broader efforts in deep learning have sought to reformulate state-of-the-art neural networks to apply to graph structures. These efforts have primarily focused on connecting graph learning to supervised and semi-supervised architectures by applying various mechanisms of neighborhood aggregation to incorporate relational information in the predictive task \cite{kipf2016semi}, \cite{velivckovic2017graph}, \cite{scarselli2008graph}, \cite{bruna2013spectral}, \cite{hamilton2017inductive}. It has become common to view many of these neighborhood aggregation algorithms as a type of message passing algorithm that transmits neighborhood information over the edges of the network\cite{Gilmer2017}\cite{GarciaDuran2017}.


While GCNs have seen significant empirical success, researchers have started to identify some of their limitations. For instance, among the known limitations of GRL algorithms is the challenge of modeling structures with hierarchies and power-law distributions, such as language as well as natural graphs. Research suggests that embedding natural graphs in hyberbolic space as opposed to Euclidean space allows for the model to effectively capture latent hierarchical structures \cite{Nickel2017}. Another recent approach applying graph coarsening via cluster assignment layers also seeks to overcome this challenge in GNN architectures\cite{Murphy2019}. Furthermore, a comparative study of a number of GRL approaches, found that certain graph properties such as clustering coefficient have a big determination on which algorithm performs better \cite{khosla2019}.

Until recently, efforts have lagged to establish a theoretical foundation for understanding GNN expressiveness, power and limitations. Primary attention in these efforts has been focused on understanding the expressiveness power of graph convolutional approaches. \cite{Maron2019}, \cite{Morris2018}, and \cite{Xu2018} define expressiveness in terms of a model's ability to test for isomorphic graphs. A common benchmark approach for testing graph isomorphism is the Weisfeler-Leman (WL) test. They prove theoretically that common GCN models are bounded in terms of their ability to distinguish graph structures by the 1-WL test. The WL test can be extended to higher-orders with increasing expressiveness. This research suggests that by modifying existing GCN architectures they can achieve up to K-WL expressiveness. The main limitation described in this research relates to the class of permutation invariant functions required to aggregate information over sets of neighboring nodes \cite{zaheer2017deepset} \cite{Xu2018}. Other recent efforts expanding beyond graph isomorphism tests follow two paths.  \cite{verma2019stability} studied the relationship between the eigenvalues of a graph convolution filter and the stability of the model. Finally, \cite{dehmamy2019} seek to quantify a representation learning algorithms expressiveness in its ability to learn graph moments. While this work was largely confined to small random graphs, it is an important precursor to the work described here.

\section{Theoretical Framework}

\subsection{Properties of Shallow and Deep Embedding Approaches}
In this section, we will explore what different GRL approaches are learning about the input graph and attempt to identify their strengths and limitations.
When it comes to the objective function, both shallow and deep embedding methods in the unsupervised setting attempt to minimize a proximity-based reconstruction error of the form:

\begin{equation}
J = - \sum_{i,j} A_{i,j} \cdot f(\phi_i, \phi_j)
\end{equation}
where $i$ and $j$ represent two nodes and $A_{i,j}$ the corresponding entry in the adjacency matrix, while $\phi$ represents the parameters to be learnt and $f$ is the proximity function. 


\subsubsection{Shallow Embedding Approaches}

Popular shallow embedding approaches extract the context around a node using co-occurence pairs in truncated random walks. At each step, the transition probability from vertex $i$ to vertex $j$ is $P_{ij} = \frac{A_{ij}}{D_{ii}}$
or in matrix notation $P = D^{-1} \cdot A$ where $D$ is the diagonal degree matrix and $A$ is the adjacency matrix.
Truncated random walks express similarity between nodes $i$ and $j$ proportional
to the sum of probabilities, denoted as $P^{t}_{ij}$, that node $j$ is reachable from a walk of length $t$ initiated from node $i$ and vice versa. 

\begin{lemma} Transition probabilities $P^{t}_{ij}$ and $P^{t}_{ji}$ through a random walk
of length t have a ratio that only depends on the degrees that are noted as $D_{ii}$ and $D_{jj}$ in the diagonal degree matrix:

\begin{equation}
    D_{ii} \cdot P^{t}_{ij} = D_{jj} \cdot P^{t}_{ji}
\end{equation}

\end{lemma}

\begin{lemma} When the length t of a random walk starting at node i tends towards infinity, the transition probability $P^{t}_{ij}$ only depends on the normalized degree of destination node j.
\begin{equation}
    \lim_{t \to \infty} P^{t}_{ij} = \frac{D_{jj}}{\sum_{k=1}^{N} D_{kk}}
\end{equation}
\end{lemma}
where $N$ is the total number of nodes in the graph.

The above hold true for the case of unbiased random walks, an example of which is DeepWalk. Node2vec, however, manipulates random walks based on the two hyperparameters $p$ and $q$ that control the probability of going backwards and the probability of favoring locality respectively. The values of these hyperparameters alter the transition probability to achieve a more balanced exploration of the neighborhood. The second order transition probability of getting from node $i$ to $j$ with the preceding node in the walk being $k$ is defined as:

\begin{equation}
    P_{i,k,j} = \begin{cases}
      \frac{1}{p} & if \ s_{kj}=0 \\
      1 & if \ s_{kj} = 1 \\
      \frac{1}{q} & if s_{kj} = 2
    \end{cases}  
\end{equation}
where $s$ stands for shortest path length.
Proofs for the above lemmas can be found in Appendix \ref{sec-appendix}.

\subsubsection{Deep Embedding Approaches}

In GRL the set of feature vectors in a node's neighborhood can be expressed as a multiset, i.e. a set with multiple entries of an element. Subsequently, aggregating neighbor features is essentially an aggregation function over multisets. The discriminative power of a node representation lies in the ability to map different multisets into different representations.
The message passing framework on which most GCN approaches in literature are based implements the aforementioned principles for equivariant transformations. Instead of external node features, these methods can also operate on identity features directly derived from the adjacency matrix. The $l_{th}$ layer in a GCN is defined as:
\begin{equation}
    h_{N(u)}^{(l)}) = AGGREGATE(h_{v}^{(l-1)}, \forall v \in N(u))
\label{eq_agg}
\end{equation}

\begin{equation}
    h_u^{(l)} = \sigma (W \cdot COMBINE(h_u^{(l-1)}, h_{N(u)}^{(l)}))
\label{eq_comb}
\end{equation}

where $u$ is the current node, $h_u^{l}$ is the feature vector of node $u$ at the $l_{th}$ layer, $N(u)$ is $v's$ neighborhood, $\sigma$ and $W$ are a non-linearity and weight matrix respectively.

Common choices for the aggregate functions are element-wise mean and max pooling as introduced in the GraphSAGE framework \cite{hamilton2017inductive}, while the combine function is often a form of concatenation with a set of linear weights followed by a non-linearity. The two functionalities can also be integrated, as happens in the GCN framework \cite{kipf2016semi}. Inherently during the aggregation step certain information about the neighborhood properties is smoothed out, since both mean and max aggregations are not strictly injective. The aforementioned aggregate functions over multisets showcase the following properties:

\begin{lemma}
Given a set $X_1$ and $X_2$ is a multiset containing fixed number of copies of the elements in $X_1$ and a function $f$ that operates on multisets, then there exists a function $h(X) = mean_{x \in X} f(x)$ such that $X_1 = set(X_2)$ and $h(X_1) = h(X_2)$.
\end{lemma}

\begin{lemma}
Given two multisets $X_1$ and $X_2$ that share the same underlying set $X_1 = X_2$ and a function $f$ that operates on multisets, then there exists a function $h(X) = max_{x \in X} f(x)$ such that $h(X_1) = h(X_2)$. 
\end{lemma}

As a direct result of Lemma 3.1, if the sets of neighborhood features contain different number of copies of the same distinct elements, i.e. neighbors that share the same feature values, then the mean aggregator will fail to distinguish between them. This aggregator captures distributions rather than the exact topological structure.
Max aggregations on the other hand simplify the multiset into its core making them robust to outliers and noise.
However, multiple nodes with the same feature are treated as a single occurence (see Lemma 3.2) therefore distributions of features inside the neighborhood set deviates from the original graph topology.
Finally, in order for these methods to scale, neighbors are often down-sampled using random sampling which fails to retain the original degree distribution. Variations on these aggregations include operating on fixed sized neighborhoods to preserve the computational footprint of the method or applying learnt weighting (e.g. with attention mechanism) in the case of mean aggregations. 

\subsection{Connection to Graph Properties}
\label{sec-graphProps}
GRL approaches differ in the process used to collect and combine the context around a node and this affects how these processes capture the complexities in the structure of a graph.
Broadly used metrics to describe these structures include communities, variations of degree, transitivity related metrics, such as clustering coefficient and distance based metrics such as closeness centrality.

{\it Community structure:} Under the homophily hypothesis embedding approaches should bring closer together highly interconnected nodes that belong to the same community.
Based on the Markovian properties of random walks, if two nodes $i$ and $j$ are in the same community, then for every intermediate node $k$ in a random walk we have transition probabilities $P_{i,k} \approx P_{j,k}$ meaning they view the rest of the network in a similar way and this is encoded in the transition probabilities. This property has been used in \cite{Pons2005Walks} to prove the correlation between random walks and community structure.
Going one step further, by tweaking the transition probabilities one can alter the extent to which community structure is represented in the resulting samples as occurs in biased random walks (e.g. Node2Vec with small $q$ parameter). It needs to be noted that sampling based GCNs that are no more than 2 layers deep observe a smoothed out version of the position of the node inside a community broader than the 2-hop neighborhood. Only when there are rich node features can the distribution of the neighborhood features provide a strong signal for community structure.

{\it Degree and Neighborhood Degree:}
As expressed in Lemma 3.3 transition probabilities are dependent on the ratio of degrees between source node and all other nodes. Therefore degree distribution is implicitly expressed in the topology sampled from a random walk, however walk length would have to approach infinity to have a direct dependence on the degree of the source (see Lemma 3.4). In biased random walks when the sampling is skewed towards locality (in-out parameter $q >1$), the variance in representing the distribution of 1-hop neighbors is reduced leading to a more accurate approximation of the source node's degree. Finally, GCN variations that operate directly on the normalized adjacency matrix should observe a more accurate degree distribution only when the sampling neighborhood size is comparable to the average degree of all the nodes in the graph. 

{\it Clustering coefficient and triangles: }
Clustering coefficient is highly dependent on reciprocity and closed triads in a graph. The count of triangles that a node participates in is an unnormalized expression of the strategic positioning of a node. For Node2Vec, if the probability of a triangle between nodes $i$, $j$ and $k$ is low then the effect of parameter $p$ is limited and biased random walks collapse into uniform ones. Similarly, in a sparse graph with clearly defined components that are disconnected with one another, parameter $q$ becomes less relevant. GCNs that operate with pooling architectures at depths higher than $l=3$ have been proven to preserve clustering coefficient \cite{hamilton2017inductive}. It is crucial though that this proof assumes that all nodes have a distinct feature representation and the network can map them to unique vectors which is not guaranteed since the aggregation functions are not injective. 

{\it Closeness Centrality:} Calculated as the reciprocal of the sum of length of all shortest paths between a node and all others, this metric is closely related with the concept of walks on a graph.
With appropriate values in parameters $p$ and $q$ Node2Vec is expected to capture the structural equivalence of nodes, hence their relative proximity to the rest of the network. A node that assumes the role of a ‘broadcaster’ in the network will be heavily represented in any form of truncated random walk generated topology. On the other hand, GCNs may find it challenging to represent high level graph structure information that cannot be easily captured by local features. In Section \ref{sec-res} we will explore the ability of GCNs to preserve such global roles of nodes.

\section{Experimental Design}
\label{sec-Exps}
\subsection{Workflow}
\label{sec-wrkfl}
Driven from the observations on the theoretical properties of GRL approaches with respect to the topology of the original graph, our goal is to design an experimental workflow to address the following research questions:
\begin{itemize}
    \item To what extent is the performance of GRL methods influenced by the structural properties of the original graph?
    \item How is the combination of input graph, embedding method and hyperparameter selection affecting performance in downstream tasks?
    \item How well are graph topological properties represented in the latent representations of GRL methods and how is this related to the hyperparameter selection?
    \item Finally, do deep GRL methods always outperform shallow ones with regards to all tasks?
\end{itemize}

\begin{figure}[h]
  \centering
  \scalebox{0.65}{
  \includegraphics[width=\linewidth]{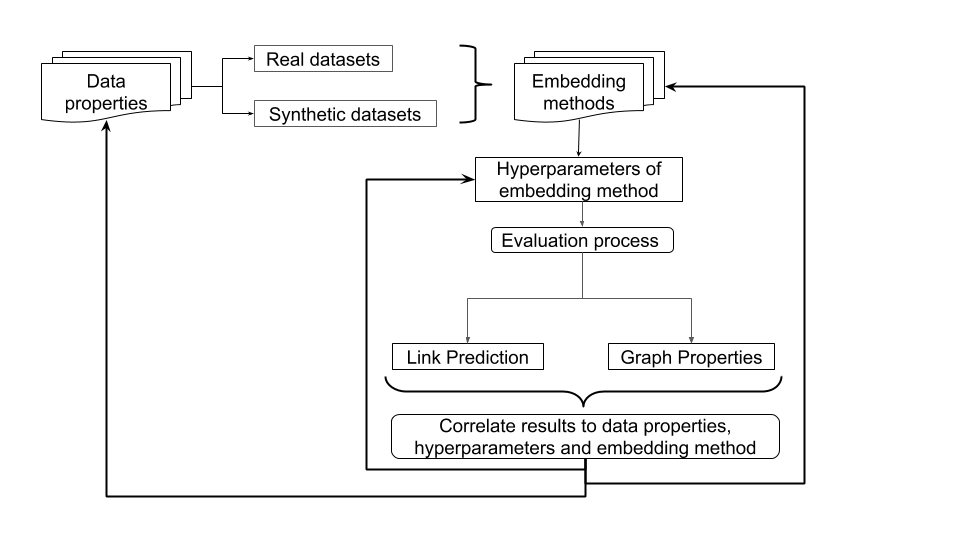}}
  \caption{Diagram of the experimental workflow.}
  \label{fig-workflow}
\end{figure}

The workflow we follow to address these questions is depicted in Figure \ref{fig-workflow}. We start with a set of structural properties that are often encountered in real world graphs and utilize both real and synthetic datasets to explore the full spectrum of these properties (Section \ref{sec-data}). We select a set of representative GRL methods and a subset of their hyperparameters that affect the data distribution the models are trained on that are described in Section \ref{sec-unsup}. For the evaluation process (Section \ref{sec-eval}) in the unsupervised case we introduce a set of learning tasks related to graph properties. Finally, we link the results back to the properties of the underlying graph and the modeling choices (Section \ref{sec-res}).

\subsection{Datasets}
\label{sec-data}
Despite the abundance of GRL approaches being introduced, most of them are still evaluated on small graphs representing a limited set of application domains (e.g. bibliographic references) that often exhibit repetitive motifs and similar values in graph statistics, such as density or transitivity. Oftentimes the underlying graph generation process of real world graphs (e.g. scale free or small-world) is associated with these graph properties. In our experiments we use a set of open source datasets widely adopted for testing in GRL literature along with a set of synthetic graphs to simulate these cases in a controlled environment and identify graph motifs that pose challenges to GRL methods. The open source datasets used are comprised of Cora and Citeseer \cite{sen2008collective}, which represent citation networks, Pubmed \cite{namata2012query} which focuses on medical publications and Reddit representing the social network built around an online blogging platform (in the version released in \cite{hamilton2017inductive}). 

\begin{figure}[]
  \centering
  \scalebox{.88}{
  \includegraphics[width=\textwidth]{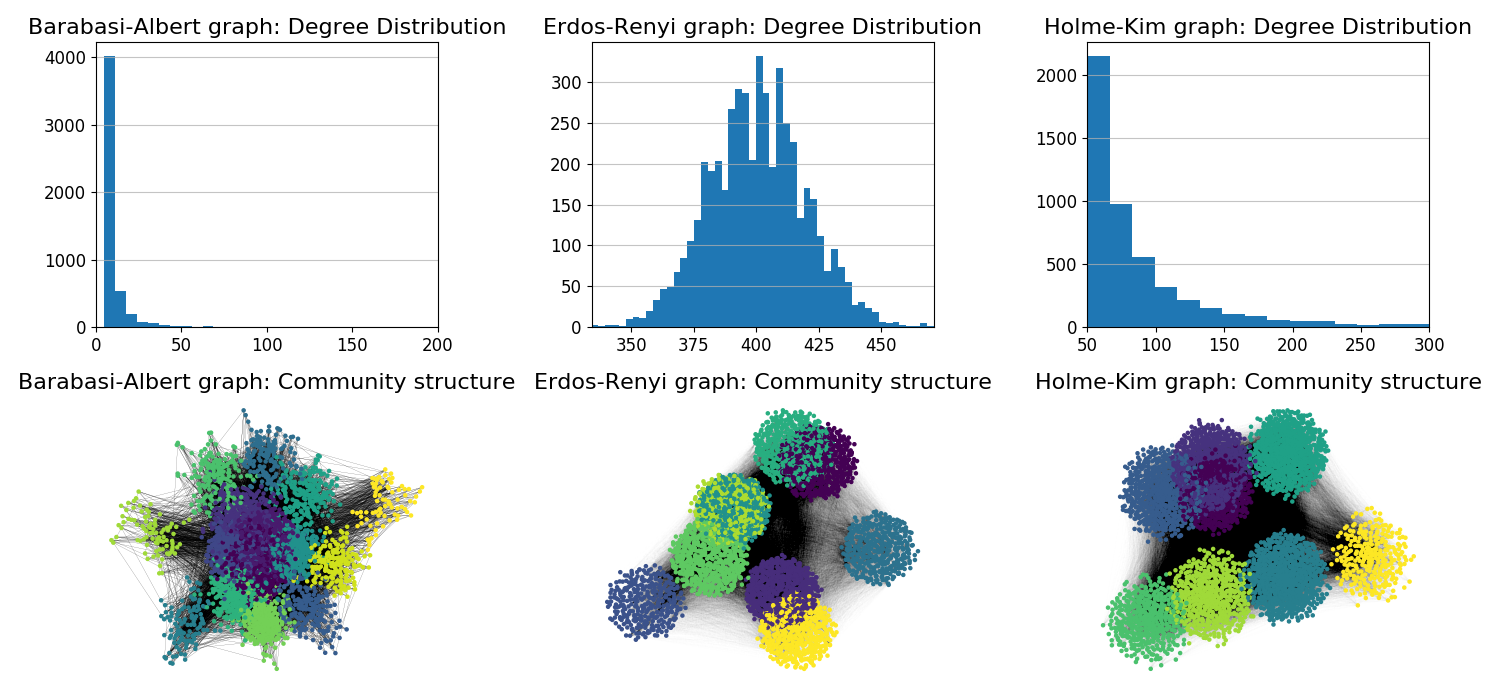}
  }
  \caption{Examples of synthetic graphs generated with the three described random graph generators: degree distributions on the first row and community structure on the second.}
   \label{fig-syntheticPlots}
\end{figure}

Many real world networks exhibit scale-free properties linked to a hierarchical topology with a small subset of nodes exhibiting a high number of connections, while the majority of the network is less connected \cite{Adcock2013TreeLikeSI}. This case is represented in our experiments with the Barabasi-Albert (BA) model with preferential attachment \cite{albert2002statistical}. In addition, many of them exhibit a small average shortest path length indicating high connectivity but also a clustering coefficient significantly higher than expected by random chance, which adheres to the Holme-Kim (HK) graph generation process \cite{holme2002growing}. Finally, the case of purely random generated graphs with a binomial edge distribution is expressed by  Erd\H{o}s-R\`enyi model \cite{newman2003random}. In practive, networks have been observed to be sparser than the purely random ones which means the average degree is smaller compared to the number of nodes $N$. In ER graphs this is achieved by setting the edge probability to a small value but maintaining $N \cdot p > 1$. This leads to an approximate Poisson degree distribution and the resulting graph is expected to have at least one large connected component. Examples of the degree distribution and community structure that arise from the different generators are shown in Figure \ref{fig-syntheticPlots}.

\begin{figure}[!htbp]
  \centering
  \scalebox{.55}{
  \includegraphics[width=\linewidth]{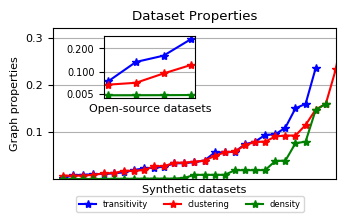}}
  \caption{Transitivity, average clustering coefficient and density values for datasets used in the experiments}
  \label{fig-syntheticStats}
\end{figure}

For our experiments we generated three different variations of BA graphs with number of edges to attach $m=2,5,9$, four ER variations with probability $p=0.002,0.004,0.008,0.016$ and finally HK graphs with two parameters: number of edges to attach ($m$) and probability of creating a triangle ($p$). The values used for HK graphs are the full set of combinations with $m=3,4,5$ and $p=0.005,0.02,0.09$. All the generated graphs contain $5000$ nodes. As can be seen in Figure \ref{fig-syntheticStats} with the addition of the synthetic graphs we are able to explore a broader range of density, transitivity and clustering coefficient values. 

\subsection{Embedding Methods}
\label{sec-unsup}
In the case of GRL methods trained in an unsupervised manner, the main learning objective is to reflect effectively the topology of the input graph either through aggregation of identity rows or through the distribution of external node features. For the experimental section we have selected popular approaches from both shallow and deep embedding methods described in Table \ref{tab-unsupMeth}. In order to explore the relationship between graph properties and GRL methods, we add one more axis of evaluation namely the {\it hyperparameters} of each method. Out of the wide range of hyperparameters we have selected the subset related to either the data distribution that gets fit into the model or the dimensionality of the resulting latent space. Hyperparameters that relate to the optimization process (such as choice of optimizer and learning rate) are out of scope for our analysis and we opt for using the default values proposed in the original implementation of each method (refer to Appendix \ref{sec-appendix}).

\begin{table}[h]
\centering
\scalebox{0.8}{
\begin{tabular}{cl}
\hline
Embedding Method           & \multicolumn{1}{c}{Hyperparameters}      \\ \hline
\multicolumn{2}{c}{Shallow methods}                                   \\ \hline
\multirow{3}{*}{DeepWalk \cite{perozzi2014deepwalk}}  & \textit{negative sampling exponent (ns)} \\ \cline{2-2} 
                           & \textit{sampling frequency (s\_freq)}    \\ \cline{2-2} 
                           & \textit{embedding dimension (d)}         \\ \hline
\multirow{3}{*}{Node2Vec \cite{grover2016node2vec}}  & \textit{return parameter (p)}            \\ \cline{2-2} 
                           & \textit{in-out parameter (q)}            \\ \cline{2-2} 
                           & \textit{embedding dimension (d)}         \\ \hline
\multicolumn{2}{c}{GCN variations}                                    \\ \hline
\multirow{4}{*}{GraphSAGE \cite{hamilton2017inductive}} & \textit{maximum degree (max\_d)}         \\ \cline{2-2} 
                           & \textit{neighborhood sample size 1 (s1)} \\ \cline{2-2} 
                           & \textit{neighborhood sample size 2 (s2)} \\ \cline{2-2} 
                           & \textit{embedding dimension (d)}         \\ \hline
\multirow{2}{*}{GCN \cite{kipf2016semi}}       & \textit{maximum degree (max\_d)}         \\ \cline{2-2} 
                           & \textit{embedding dimension (d)}         \\ \hline
\multirow{2}{*}{Graph Attention Network (GAT) \cite{velivckovic2017graph}}       & \textit{number of heads (n\_a)}          \\ \cline{2-2} 
                           & \textit{embedding dimension (d)}         \\ \hline
\end{tabular}
}
\caption{Description of embedding methods and their hyperparameters used in the experimental process}
\label{tab-unsupMeth}
\end{table}

\subsection{Evaluation}
\label{sec-eval}
To address the research questions posed in Section \ref{sec-wrkfl} about the relationship between the properties of the original graph and the performance of the embedding methods, we expand the evaluation criteria beyond link reconstruction. We define predictive models for the node-level topological graph properties discussed in Section \ref{sec-graphProps}.
The goal of this evaluation is to explore which topological properties are preserved in the latent space by each GRL method and what part of this connection can be traced back to the type of generation process that the graph adheres to. The aforementioned node level properties can characterize roles inside a network that are crucial for downstream learning tasks. More specifically, a post hoc supervised predictor is fit to predict each of these properties using the embedding vector of each node as input. For the degree and average neighborhood degree predictor we log-transform the target variable and for community memberships the class assignment is based on Louvain community detection \cite{blondel2008fast}. We use Random Forest predictors for this task with the number of estimators ranging from 100 to 500 and the depth of each tree from 8 to 12. A split of 90/10 for train and test set is employed and the best performing results are reported for each predictor in Section \ref{sec-res}. However, the hyperparameters for the Random Forest models remain the same for a given dataset across results derived from different methods to avoid biasing the evaluation of a method due to more efficient hyperparameter tuning. Regarding model selection we opted for a non linear model since the association between the latent space and the target metrics is unlikely to follow linear correlation. Opting for Random Forest models allows for minimal hyperparameter tuning on the downstream model to avoid biasing the subsequent sensitivity analysis. With six GRL methods and on average ten hyperparameter per method and over 20 datasets, the total scale of our sensitivity analysis is over 1000 GRL models trained in total. Accounting for the models trained for the prediction of graph properties as part of the evaluation we reach a total of 8000 models constituting this the largest sensitivity analysis in the field of graph representation learning. 

\section{Results}
\label{sec-res}

\begin{figure}[h]
  \begin{center}
  \begin{adjustbox}{minipage=\textwidth,scale=0.75}
  \begin{subfigure}{\textwidth}
  \includegraphics[width=\textwidth]{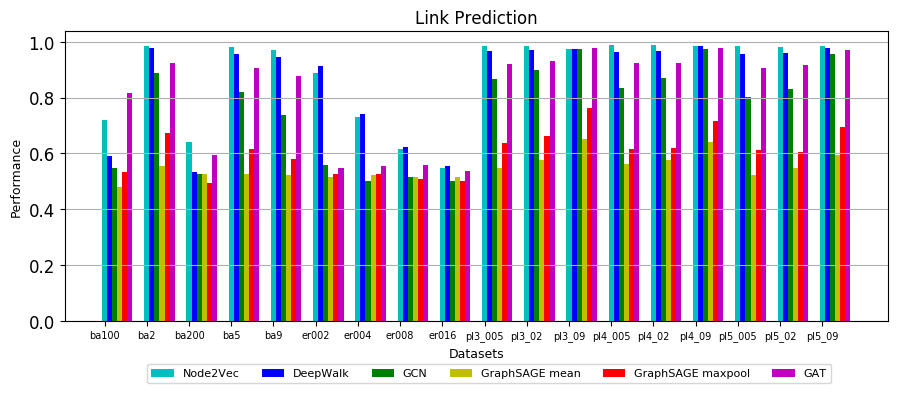}
  \caption{Synthetic datasets}
  \end{subfigure}
  \begin{subfigure}{\textwidth}
  \includegraphics[width=\textwidth]{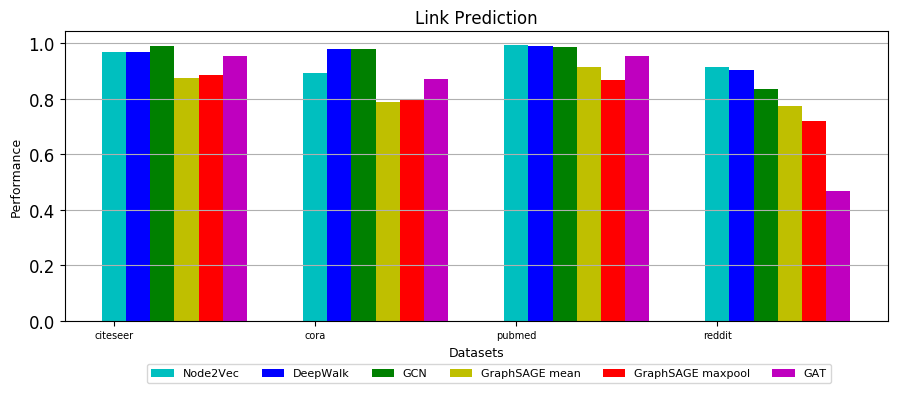}
  \caption{Open Source Datasets}
  \end{subfigure}
 \end{adjustbox}
 \end{center}
 \caption{Link prediction AUC scores for different datasets and GRL methods}
  \label{fig-LPbar}
\end{figure}

The most popular evaluation task used for GRL methods that are not trained to optimize node class assignment is {\it link prediction} which is depicted in Figure \ref{fig-LPbar}. Shallow methods seem to outperform deep GRL approaches across almost all graphs. Only GAT appears to be on par with the shallow embedding methods, which implies that applying attention weights over a node's neighborhood can significantly improve the reconstructive power of the produced embeddings. DeepWalk and Node2Vec are also able to adapt better to scale-free graphs and heavily clustered graphs ({\it BA} and {\it HK}). In highly dense random graphs ({\it ER}) all GRL seem to be performing poorly, which is to be explained since in the random edge creation process there is wider variety of connectivity patterns to learn. For the Reddit dataset, the largest of the contemplated graphs, GCN variations appear to suffer significantly in performance which can be explained by the sampling of neighborhoods that needs to occur for these methods to effectively scale.

\begin{figure}
  \centering
  \scalebox{0.95}{
  \includegraphics[width=\textwidth,width=\textwidth]{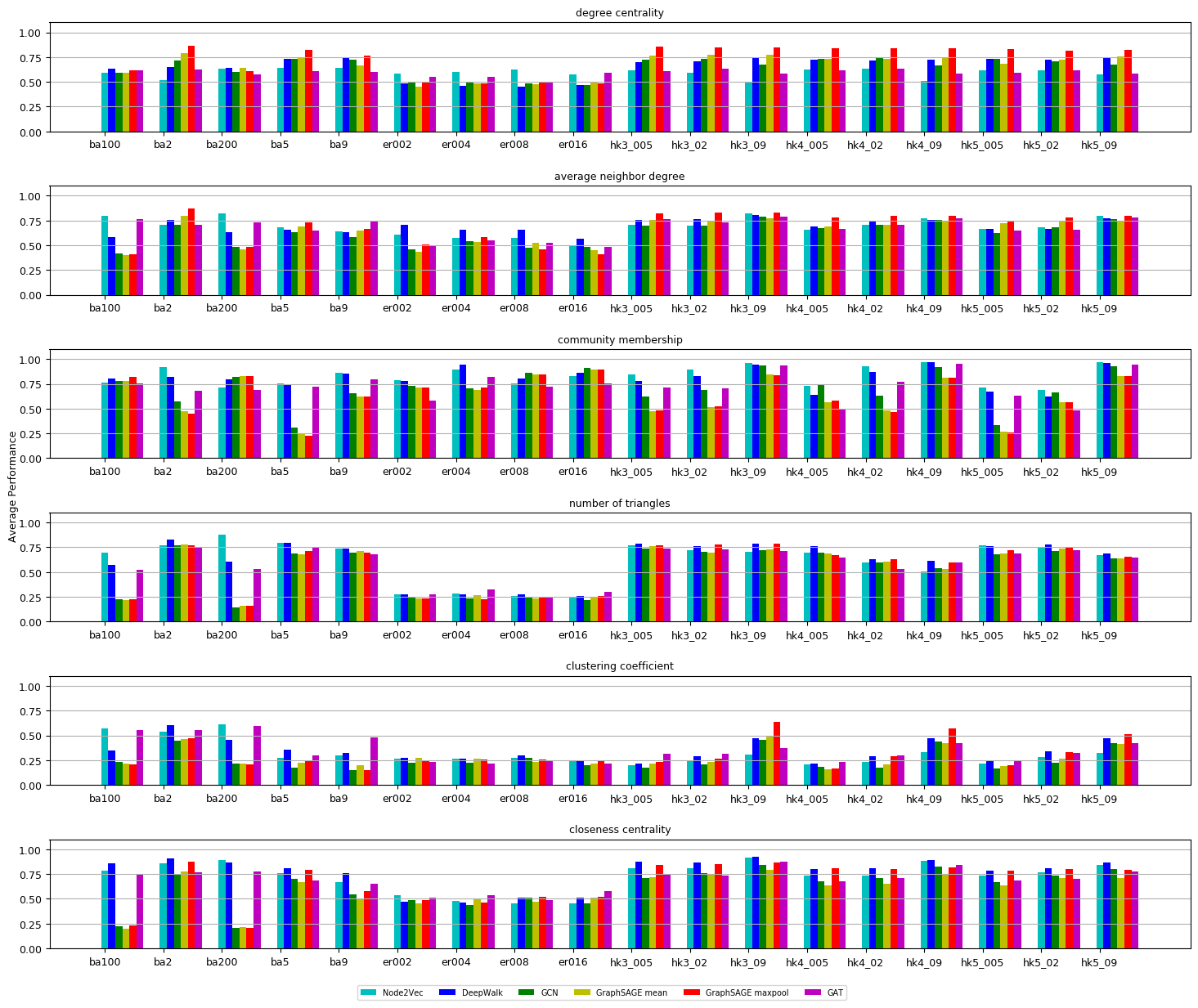}
  }
  \caption{Performance of different GRL methods for six graph properties prediction tasks for synthetic graphs.}
  \label{fig-indGPSynth}
\end{figure}

\begin{figure}
  \centering
  \scalebox{0.95}{
  \includegraphics[width=\textwidth]{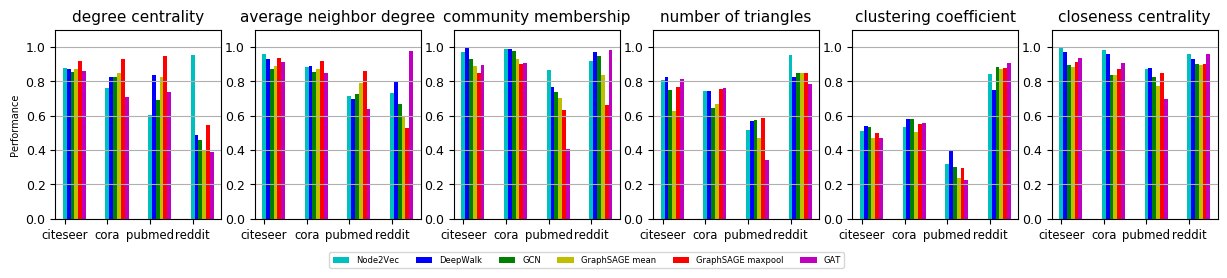}
  }
  \caption{Performance of different GRL methods for six graph properties prediction tasks for open source graphs.}
  \label{fig-indGPOpen}
\end{figure}

Figures \ref{fig-indGPSynth} and \ref{fig-indGPOpen} demonstrate performance across all graph properties prediction tasks for different GRL methods corresponding to the best performing hyperparameter combination. Performance is expressed in terms of $Micro-F1$ and $R^2$ scores for classification and regression tasks respectively. We identify a trend of GCN variations managing to preserve individual node degree and average neighborhood degree as opposed to shallow embedding methods that show a higher success rate in capturing community structure and closeness centrality. In the case of synthetic graphs in particular where there is no rich feature set to accompany the nodes, transition probabilities used in shallow methods encode nodes belonging in the same community more effectively as explained in Section \ref{sec-graphProps}. Derived from Lemma 3.3, however, finite random walks capture ratios of degree with an advantage to biased walks with increased locality sampling. This is evident by the good performance of Node2Vec in predicting degree and the overall better performance of GCNs in degree related tasks as they have a more complete view of 1- and 2-hop neighborhoods. Node2Vec appears to be performing well in high reciprocity graphs (e.g. BA) in tasks related to capturing closed triads. As explained is Section \ref{sec-graphProps}, random walks efficiently capture 'broadcaster' node behavior and only addition of attention weights improves the discriminative power of GCNs in this matter. Performance is significantly higher across tasks for the simpler open source graphs which adheres to our original observation that existing benchmarks are not providing enough variation. In particular for graphs with hierarchical properties like {\it BA} and {\it HK} where there exist supernodes that cause community overlap (refer to Figure \ref{fig-syntheticPlots}) the difference in performance between embedding methods is even higher for the community identification task.

\begin{figure}[!ht]
\begin{subfigure}{\linewidth}
  \centering
  \includegraphics[width=\linewidth]{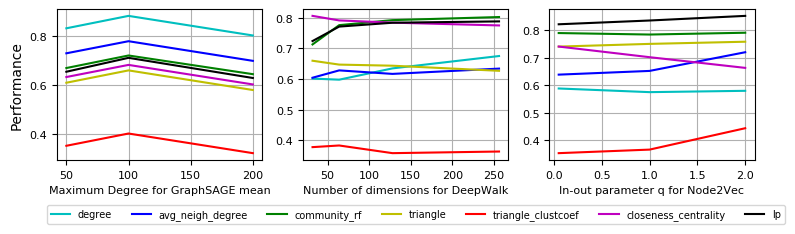}
  \caption{Barabasi-Albert graphs}
  \label{fig-ba_avgHyp}
\end{subfigure}
\begin{subfigure}{\linewidth}
  \centering
  \includegraphics[width=\linewidth]{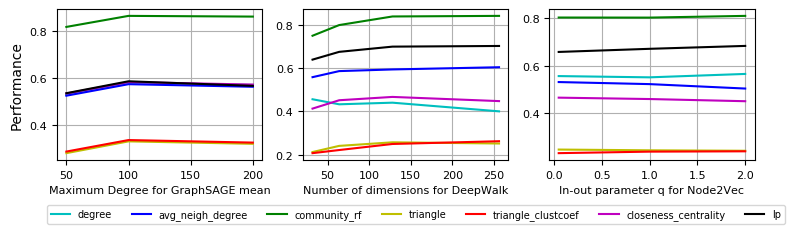}
  \caption{Erd\H{o}s-R\`enyi graphs}
  \label{fig-er_avgHyp}
\end{subfigure}
\begin{subfigure}{\linewidth}
  \centering
  \includegraphics[width=\linewidth]{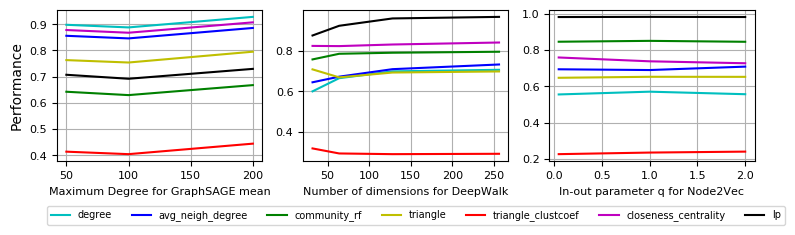}
  \caption{Holme-Kim graphs}
  \label{fig-pl_avgHyp}
\end{subfigure}
\caption{Performance across individual evaluation tasks for the three categories of synthetic graphs and different hyperparameters to method combinations.}
\label{fig-synth_avgHyp}
\end{figure}

Next, we explore the effect of individual hyperparameters on the performance in different evaluation tasks. In Figure \ref{fig-synth_avgHyp} examples of hyperparameter values are plotted against performance in a subset of tasks averaged over all instances of a given graph generator. Interestingly, BA graphs appear to be more affected from hyperparameter values and different graph generators respond differently to a given hyperparameter setting which indicates that sensitivity analysis of GRL methods needs to include multiple types of graphs. More specifically, it appears that the maximum degree value selected to subset each node's neighborhood in GraphSAGE with mean aggregator is related with increased performance as the degree value is set higher for HK graphs. The other two graph types plateau or even drop in performance when the value is increased above 150. This can be attributed to the thicker tail in degree distribution for HK graphs (see Figure \ref{fig-syntheticPlots}) with more nodes displaying high degree values. An interesting observation is the effect of parameter $q$ for BA graphs, with higher $q$ value increasing performance for clustering coefficient and average neighborhood degree, while it drops for closeness centrality prediction. This showcases the difference between local and global structure, since parameter $q$ regulates the exploration of immediate neighbors and the ones at a higher depth. Finally, link prediction in BA graphs achieves best performance for smaller number of dimensions but all other tasks require collectively higher dimensionality. This evidence suggests that there is not necessarily a single value in hyperparameter selection that achieves optimal performance for all evaluation tasks.

\begin{figure}[!ht]
  \centering
  \scalebox{0.92}{
  \includegraphics[width=\linewidth]{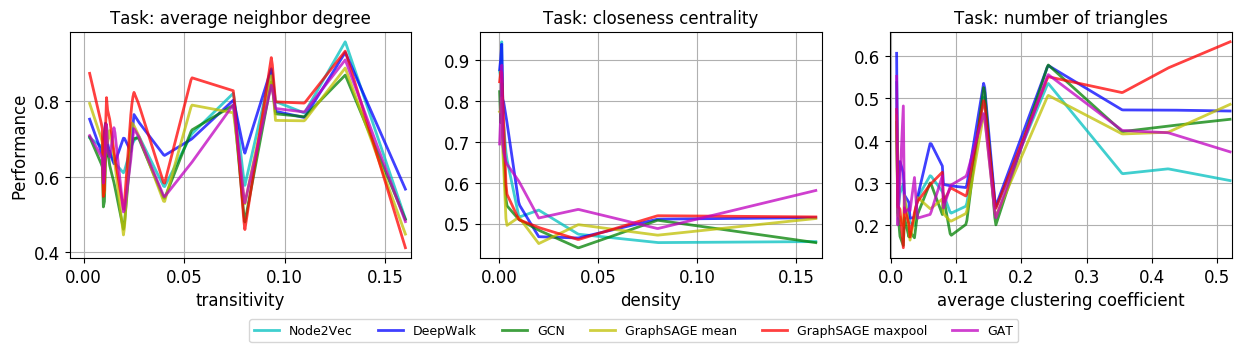}
  }
  \caption{Relationship of performance of individual evaluation tasks to increasing values of transitivity, density and average clustering coefficient.}
  \label{fig-grProps}
\end{figure}

Finally, we attempt to quantify the relationship between the features of the input graph to the performance in each of the evaluation tasks, as depicted in Figure \ref{fig-grProps}. It needs to be noted that this relationship is not always smooth, since the different values on graph features could also correspond to a different graph generation process. We opted for this abstraction to correlate a single graph property to an evaluation task irrespective of other particularities of the graph. In this comparison, GraphSAGE maxpool appears to be outperforming all other approaches for lower value of transitivity, while it faces a steep drop in performance in higher transitivity cases. As the density increases, the "commute time" to get from one node to another increases and higher order walks would be needed to accurately reflect the strategic placement of a node, i.e. closeness centrality, which causes performance of shallow methods to drop.
Finally, high average clustering coefficient (e.g. in the case of HK graphs) indicates that across the network a node's neighbors are also neighbors of each other therefore an aggregation of neighbor's features serves as a good representation for a node in the latent space. 


{\bf Practitioner's guidelines.} Our evaluation process provided evidence that there is not a "one-to-fit-all" approach in the application of GRL. Different types of graphs respond differently to a hyperparameter setting which indicates that a sensitivity analysis of a GRL method needs to be specialized to the type of input graph. Suggested default values are often derived from graphs with different properties and thus cannot be easily transferred to a new application. It appears that GRL approaches do not adapt equally well to the particularities of the input graph which can be problematic in a real world application scenario. Dependent on the downstream learning task, whether that means identifying strategic nodes or the ones that bridge together different communities, the selection of GRL method may vary. As with other graph mining tasks, capturing both local and global structure is challenging and we provide evidence that popular GRL approaches often manage to preserve one of the two. Random walk based methods appear to capture global roles, e.g. community memberships, efficiently and adapt to heavily clustered graphs with power law degree distributions whereas GCNs capture local properties, such as degree, and the application of attention weights on them provides a significant boost. Our evidence suggests that graphs with Poisson degree distribution (ER) respond better to shallow embedding approaches. BA graphs are well combined with both shallow approaches and attention GCNs. GraphSAGE with max pooling appears well suited for graphs with hierarchical structure but high local clustering, i.e. HK graphs.
Our analysis indicates that the choice of embedding algorithm on each own is not the key decision point in the application of GRL. Instead, the identification of the characteristics of the input graph and the definition of the properties that need to be preserved for downstream learning tasks are the decisive factors that determine the applicability of GRL in a real world scenario. Following such an evaluation process improves interpretability in the application of GRL and provides verification for the choice of method and hyperparameter setting based on application needs and properties of the input graph.

\section{Conclusions}
In the present work we propose an extensive data-intensive evaluation approach of GRL methods with respect to the characteristics of the input graph, the hyperparameters of each method and the properties that are of importance for downstream learning applications hence need to be preserved in the latent space. We provide evidence on test cases that challenge each family of methods and distill applicability guidelines for practitioners in the process to select the appropriate GRL method for their purposes.

\appendix

\section{Supplementary Material}
\label{sec-appendix}

In the experimental section the following hyperparameters are contemplated: maximum degree $\in [50,100,200]$ for GraphSAGE \footnote{https://github.com/williamleif/GraphSAGE} and GCN \footnote{https://github.com/tkipf/gcn}, neighborhood sample size at 1-hop $\in [25,50,100]$ and at the 2-hop neighborhood sample size $\in [10,20,40]$ for GraphSAGE, number of attention heads $\in [4,16,32]$ for GAT \footnote{https://github.com/PetarV-/GAT}, parameter $q$ $\in [0.5,1,2]$ and parameter $p$ $\in [1,2]$ for Node2Vec \footnote{https://github.com/aditya-grover/node2vec}, subsampling frequency $\in [0.0001,0.001,0.01]$ and negative sampling exponent $\in [-0.75,0,0.75]$ for DeepWalk \footnote{https://github.com/phanein/deepwalk} and finally embedding dimensionality $\in [32,64,128,256]$. The remaining hyperparameters of each method follow the default values proposed in the original implementations of each method. It needs to be noted that for a subset of the remaining hyperparameters, namely number and length of walks and number of training epochs a separate grid search was carried out for each dataset while maintaining the contemplated hyperparameters at default values and the best performing ones were maintained throughout the subsequent sensitivity analysis. The values explored during this grid search include number of walks $\in [5,8,10,32,64]$, walk length $\in [5,25,40,80]$ and epochs $\in [1,2,3,4,5,8,10]$. The selected values are subsequently maintained constant throughout the experiments for a given dataset and method combination. The hyperparameters that were contemplated as part of our sensitivity analysis were selected due to the fact that they can alter the way information from the graph gets aggregated as part of the embedding training. 

For the predictive models of graph properties in the evaluation part of the workflow we perform a 90/10 train-test split. As for the embedding training we perform edge level split with the same ratio. It needs to be noded that the inductive approaches, such as GraphSAGE, are built are proposed to operate on a node level split, meaning a subset of nodes is wihtheld and no edges related to them are seen during training. To perform fair comparisons with transductive approaches we adjust the training process of these methods to include edge-level split. 

All GRL methods are implemented with Tensorflow 1.14.0 as backend except for the GAT implementation that operates on PyTorch 1.1.0. For the graph convolutional approaches GPU optimized implementations are used. It needs to be noted that for the GAT network there was not a significant lift in performance using the GPU implementation for some of the denser graphs we contemplated. The following table contains information on the open source datasets we used in the experiments:

\begin{table}[!htbp]
\centering
\begin{tabular}{lcc}
\hline
{\bf Dataset}  & {\bf Number of nodes} & {\bf Number of edges} \\ \hline
Cora     & 2,708           & 5,278           \\ \hline
Citeseer & 3,327           & 4,676           \\ \hline
Pubmed   & 19,717          & 44,327          \\ \hline
Reddit   & 231,443         & 11,606,919      \\ \hline
\end{tabular}
\caption{Statistics of open source datasets used in the experiments}
\end{table}

For the case of Reddit, following the experimentation setting described in \cite{hamilton2017inductive} we prune out disconnected nodes and preserve the largest connected component with $168,730$ nodes.

\subsection{Proof of Lemma 3.1}
Lemma 3.1 can be written in matricial notation as $DP^{t}=D(P^{t})^{T}$ and given that $P=D^{-1}A$ we have:
\begin{equation}
    DP^{t} = D(D^{-1}A)^{t} = D (A^{T}(D^{-1})^T)^{t} = D(P^{t})^{T}
\end{equation}

\subsection{Proof of Lemma 3.2}
The transition probability matrix $P$ shares eigenvalues with its normalized version $S=D^{\frac{1}{2}}PD^{-\frac{1}{2}}$ and because it is stochastic, i.e. transition probabilities originating from the same source sum up to 1, its largest eigenvalue will be $\lambda_{1} = 1$ and for all other eigenvalues $1>\lambda_{m}>-1$. If $v_m$ and $u_m$ are the right and left eigenvectors corresponding to $\lambda_{m}$, 
the spectral decomposition of transition matrix $P$ can be written as: 
$P^{t} = \sum_{m=1}^{N} \lambda_{m}^{t} v_m u_m^{T}$.
Individual transition probabilities are then written as: 
$P^{t}_{ij}=\sum_{m=1}^{N} \lambda_{m}^{t} v_m(i) u_m(j)$.
When t tends to infinity, all the terms $m>=2$ disappear and by normalizing the eigenvectors we get:
$v_1=\frac{1}{\sum^{N}_{k=1} D_{kk}}$ and 
$u_1=\frac{D_{jj}}{\sum^{N}_{k=1} D_{kk}}$.
Therefore we have:
 
 \begin{equation}
     \lim_{t \to \infty} P^{t}_{ij} = \lim_{t \to \infty} \sum_{k} \lambda_l^{t} v_l u_l^{T} = v_1(i)u_1(j)= \frac{D_{jj}}{\sum^{N}_{k=1} D_{kk}}
 \end{equation}

\bibliographystyle{unsrt}  
\bibliography{grl}  


\end{document}